\documentclass[final,5p,times,twocolumn]{elsarticle}

 \usepackage[numbers]{natbib}

\usepackage{amsthm,amssymb,mathrsfs,setspace,pstricks,booktabs,mathtools,amsmath}
\usepackage{color, colortbl}
\usepackage{subcaption}
\usepackage{algorithm}
\usepackage{algpseudocode}
\usepackage{url}
\usepackage[normalem]{ulem}
\definecolor{LightCyan}{rgb}{0.88,1,1}
\begin{document}
\let\WriteBookmarks\relax
\def\floatpagepagefraction{1}
\def\textpagefraction{.001}



\title{A clustering and graph deep learning-based framework   for COVID-19 drug repurposing}                       

\author%
[3,2,1]{Chaarvi Bansal \corref{cor1}} 
\ead{f20180913@pilani.bits-pilani.ac.in} 
\author[3]{Rohitash Chandra \corref{cor1} } 
\ead{rohitash.chandra@unsw.edu.au} 
\author[2]{Vinti Agarwal}    
\author[1]{P. R. Deepa} 

\address[3]{Transitional Artificial Intelligence Research Group, School of Mathematics and Statistics, UNSW Sydney, Sydney, Australia}
\address[1]{Department of Biological Sciences, Birla Institute of Technology and Science Pilani, Rajasthan, India}
\address[2]{Department of Computer Science and Information Systems, Birla Institute of Technology and Science Pilani, Rajasthan, India} 

\cortext[cor1]{Corresponding author}


 \journal{*}

\begin{abstract} 
Drug  repurposing (or repositioning) is the process of finding new therapeutic uses for drugs already approved by drug regulatory authorities (e.g., the Food and Drug Administration (FDA) and Therapeutic Goods Administration (TGA)) for other diseases. This involves analyzing the interactions between different biological entities, such as drug targets (genes/proteins and biological pathways) and drug properties, to discover novel drug-target or drug-disease relations.  
Artificial intelligence methods such as machine learning and deep learning have successfully analyzed complex heterogeneous data in the biomedical domain and have also been used for drug repurposing. This study presents a novel unsupervised machine learning framework that utilizes a graph-based autoencoder for multi-feature type clustering on heterogeneous drug data. The dataset  consists of 438 drugs, of which 224 are under clinical trials for COVID-19 (category A). The rest are systematically filtered to ensure the safety and efficacy of the treatment (category B). The framework solely relies on reported drug data, including its pharmacological properties, chemical/physical properties, interaction with the host, and efficacy in different publicly available COVID-19 assays. Our machine-learning framework reveals three clusters of interest and provides recommendations featuring  the top 15 drugs for COVID-19 drug repurposing, which  were shortlisted based on the predicted clusters that were dominated by category A drugs. The anti-COVID efficacy of the drugs should be verified by experimental studies. Our framework  can be extended to support other datasets and drug repurposing studies, given open-source code and data availability. 

\end{abstract}




\maketitle
 
\section{Introduction}
Coronavirus disease 2019 (COVID-19), caused by SARS-CoV-2,  belongs to a large family of coronaviruses that are known to cause respiratory illnesses \cite{wu2020sars}. COVID-19  was declared a pandemic in March 2020 by the World Health Organisation \cite{cucinotta2020declares}, and it has deeply impacted the world economy and burdened the healthcare systems globally. \textcolor{black}{ As of 16th March 2023, COVID-19 has caused more than 681 million infections worldwide, with more than 6.81 million deaths \cite{dong2020interactive}}. The COVID-19 outbreak has presented the scientific community with unique and complex challenges. Viral complications such as multi-organ failure, septic shock, respiratory failure, and cardiovascular complications have been identified as the most common cause of death \cite{elezkurtaj2021causes,perkins2022effect}. In addition to this, many patients have also experienced long-term health problems such as fatigue, brain fog, and breathing difficulties  that have lasted months after the acute viral infection phase was over \cite{sudre2021attributes}.

The development of different types of vaccines such as messenger ribonucleic acid (mRNA)  vaccines (Pfizer \cite{polack2020safety} and Moderna \cite{baden2020efficacy}), viral vector vaccines (Johnson \& Johnson’s Janssen \cite{sadoff2021safety},     AstraZeneca-Oxford COVID-19 vaccine \cite{voysey2021safety}), and inactivated virus-based vaccines (Covaxin \cite{sapkal2021inactivated}), has played a significant role in reducing transmission rates, hospitalization rates, and disease severity \cite{vacInfo}. However, none of these vaccines have proven to be fully effective against  emerging COVID-19 variants \cite{Infovac}. Thus, there is a dire need to  develop other alternatives to reduce the impact of COVID-19 on health and society. We note that the major  goal of vaccination   was to limit virus transmission and build immunity against future infections. It has been challenging to develop/identify drugs for treating patients infected with COVID-19 irrespective of their vaccination status.
 
Drug discovery is a complex, time-consuming, and costly process with  a high degree of uncertainty about the drug's effectiveness \cite{drugDisc}. On average, the cost of developing a new drug ranges from \$314 million to \$2.8 billion, and late-stage failure of a drug is widespread \cite{jama.2020.1166}. Drug repurposing (also known as  therapy switching and drug repositioning) is an effective strategy to   get effective and affordable treatment alternatives in  a reasonable time frame. It involves identifying new therapeutic purposes for approved drugs in the market \cite{sahoo2021drug} by regulatory authorities such as the Food and Drug Administration (FDA) \footnote{\url{https://www.fda.gov/}} and the Therapeutic Goods Administration (TGA) \footnote{\url{https://www.tga.gov.au/}}. Certain antimalarial drugs such as Hydroxychloroquine, Chloroquine and antiviral drugs such as  Remdesivir, Favipiravir, and Molnupiravir have been repurposed for the treatment of COVID-19 \cite{sahoo2021drug}. However, till now, only Remdesivir has been approved by FDA for COVID-19 \cite{remdisvir}. Hydroxychloroquine was proven ineffective against COVID-19 as it had little to no impact on lowering death risk \cite{Singh21}. Similarly, Molnupiravir was found to be mutagenic to mammalian cells and hence was deemed unsafe for COVID-19 treatment \cite{zhou2021beta}. Thus, there is a need to identify more drugs or combinations of drugs that can be repurposed for COVID-19 treatment. 
 
Artificial intelligence methods such as machine learning and deep learning have been prominent in analyzing complex  data in the biomedical domain and have been used in various ways for COVID-19 \cite{dogan2021systematic} that includes modelling infections \cite{car2020modeling}, and also studying the behaviour of the population during different phases of the pandemic in social media \cite{chandra2021covid}. Machine learning methods   have also been used for  drug repurposing. Several studies have modelled the problem of drug repurposing as a classification task and employ various methods such as  support vector machines (SVM) \cite{suykens2009support}, k-nearest neighbours (KNN) \cite{shen2003development}, and random forest classifiers (RFC) \cite{susnow2003use}. By predicting the interaction between drug-target or between drug-disease, these studies have reduced the number of redundant compounds. Similarly, previous studies \cite{liu2022ai} have also developed robust techniques for COVID-19 drug repurposing by using deep learning algorithms. Since these algorithms are adept in finding underlying patterns in large unstructured data, these are widely used to solve problems involving large-scale biological network data, which is otherwise cumbersome to analyze and interpret \cite{muzio2021biological}. Some of the deep learning methods that have been used include  convolution neural networks (CNNs) \cite{hooshmand2021multimodal},  and recurrent neural networks (RNNs) such as the  long short-term memory (LSTM) network \cite{zhang2021deep,lee2021new}. Various studies have shown that approaches based on network analysis are competitive strategies for drug repurposing as they can effectively model the nature of the interaction between various biological entities, such as genes, proteins, targets, and drug interactions. \cite{ahmed2022comprehensive}. We can check for the safety and efficacy of a drug by analyzing the relationships in the pharmacological network of a drug. These pharmacological networks are large, dense networks that are heterogeneous. With the recent advancements in machine learning, the analysis of these networks can be automated to achieve results of high-quality \cite{hameed2018two}. 
 
Graph-based  machine learning and deep learning techniques have gained momentum in the past few years due to the extensive expressive power of graphs. Graphs can effectively represent complex networks such as social media, protein networks, gene networks, and publication networks \cite{zhou2020graph}. Graph neural networks have been used extensively in bioinformatics and computational biology  to process and analyze dense biological network data, including gene-gene interaction, gene-protein interaction, drug-target interaction, etc. \cite{zhang2021graph}. Drug repurposing studies utilizing graph machine learning integrate information about viral-host interaction, drug-protein interactions, drug-target interactions, and drug-viral interactions by creating heterogeneous knowledge graphs to predict drug candidates for COVID-19 drug repurposing. For instance, Hsieh et al. \cite{hsieh2020drug} derived drug embeddings from drug knowledge graphs and used Bayesian pairwise ranking and clinical trial drug embeddings to rank the top 22 drug candidates for COVID-19 drug repurposing. Saleem et al. \cite{al2021knowledge} developed a biomedical knowledge graph model and a  ranking algorithm to predict the top 50 performing compounds and conducted a molecular functional analysis to suggest drugs for repurposing. A variety of studies have also used different graph neural network architectures to extract enriched drug embeddings for link-prediction tasks or downstream node classification tasks. Ioannidis et al. \cite{ioannidis2020few}, for instance, extended relational graph convolution network (RGCN) and developed   inductive-RGCN (I-RGCN) to learn node and edge embeddings for few-shot link prediction to predict and rank candidates for COVID-19 treatment. Boutorh et al. \cite{boutorh2022graph}, on the other hand, used a graph neural network (GNN) based GraphSAGE \cite{hamilton2017inductive} model to extract unsupervised node embeddings from the knowledge graph for link prediction and subsequent ranking of drugs. 
 
Data availability is a major challenge that is faced by most drug repurposing studies.  A plethora of open-sourced databases are available such as PubChem\footnote{https://pubchem.ncbi.nlm.nih.gov/}, Clinical trials.gov\footnote{https://clinicaltrials.gov/}, Drugs@FDA\footnote{https://www.accessdata.fda.gov/scripts/cder/daf/index.cfm}, KEGG database\footnote{https://www.kegg.jp/kegg/}, and Drug bank\footnote{https://go.drugbank.com/} \cite{macraild2022systematic}; however, these present data in an unstructured format thus making it difficult to interpret. In order to address this problem, an open-source  web application known as CoviRx \cite{hardik2022systematic}  was developed  that gave access to drug data for more than 7000 drugs as on 10th November 2022. In a previous study, data on CoviRx was used to manually downselect 7000 drugs to the top 12 performing compounds, which were then selected for wet lab experiments involving human organoid models \cite{macraild2022systematic}. Organoid models   derived from human stem cells   are typically used for studying genetic and infectious diseases. They have been extensively used in cancer research and often complement animal models. Since organoid models provide a good representation of human physiology, they can also be used to study the efficacy of drugs against diseases \cite{kim2020human}. Machine learning has not yet been applied to data from organoid models to predict top-performing drug candidates for COVID-19 drug repurposing. We note that the CoviRx drug dataset typically consists of drug features of alpha-numeric values, such as gene targets, pathways, and pharmacokinetic/pharmacodynamic (PK/PD), which are difficult to process and interpret by traditional machine learning techniques. Hence, sophisticated  machine learning methodologies need to be used  to extract the non-linear relationships among these different drug properties and effectively combine them to yield top-performing candidates for repurposing. Therefore, the graph-based representation will be most effective for encoding the drugs into a machine-learning framework. \textcolor{black}{Furthermore, we note that the data is unlabelled; hence, we need to use unsupervised machine learning methods to group them. Grouping such data can enable a better understanding of drugs due to their similarities, and hence it would guide in down-selecting them for drug re-purposing. }
 
In this study, we develop an unsupervised machine-learning framework that features graph machine learning and clustering to recommend  drugs for COVID-19  drug repurposing that experimental studies can evaluate. The major advantage of the framework is that it shortens the list of drug candidates for clinical trials. The framework achieves this by combining data from a few prospective drug candidates that were down-selected for COVID-19 drug repurposing  \cite{macraild2022systematic} and drugs that have undergone COVID-19 clinical trials. Our main objective is to capture and utilize the non-linear relationship between various drug features such as biological, pharmaceutical, and chemical properties to suggest drugs for COVID-19 drug repurposing by clustering similar drugs together. The framework proposed in this study will help accelerate COVID-19 drug repurposing research and form a basis for future drug repurposing studies. \textcolor{black}{The focus of our study is the development of a machine learning framework that can be used for different variants. The framework recommends drugs that can be tested for current and emerging variants. The framework has been designed keeping in mind the possibility of new infections in the future and thus can be easily fine-tuned for other datasets.} We also provide an open-sourced python implementation of the framework to extend the research of COVID-19 drug repurposing further.

\section{Methodology}

\subsection{Source of Data}
 
Due to a lack of open-source and standardized datasets for COVID-19 drug repurposing, MacRaild et al. \cite{macraild2022systematic} manually assembled the drug data for over 7000 drugs from various datasets such as drug bank, which was then made readily available for the public via the CoviRx website developed by Jain et al. \cite{hardik2022systematic}.  

CoviRx is  an interactive website that contains data for  compounds for COVID-19 drug repurposing. CoviRx is a one-stop-shop solution for accessing data related to different features of a drug. It lists the drug's various pharmaceutical and chemical properties, risks associated with the drug, original indication data, clinical trial information for COVID-19, safety, biological data such as target, pathway, etc., pharmacokinetics, and COVID-19 assay data points \cite{macraild2022systematic}. As of June 2022, 7817 compounds were available in Compounds Australia's Open Drug Collection\footnote{\url{https://www.griffith.edu.au/griffith-sciences/compounds-australia}}, and these can be readily sourced for evaluation in CoviRx \cite{simpson2014overview}.  CoviRx data has been collected from a wide variety of sources such as PubChem, Clinical trials, Drugs@FDA, KEGG database, Drug bank, etc. \cite{macraild2022systematic}. 


The process of drug downselection results in a small manageable subset of drugs that can potentially offer safe, effective, and affordable treatment alternatives. These drugs can then be subjected to rigorous experimental studies to evaluate their efficacy.   In a previous study \cite{macraild2022systematic}, manual down-selection was done   which is feasible for smaller datasets. However, as the datasets become larger, automation becomes imperative for effective drug screening and down-selection. Therefore, in this study, we develop a machine-learning model to suggest drugs for COVID-19 drug repurposing. In our study, the machine learning framework   implements  multi-feature type clustering on a set of 438 drugs selected from 7817 drugs present on the CoviRx platform to suggest starting points for COVID-19 drug repurposing. First, we applied an assay filter and FDA/TGA approval filter similar to Macraild et al. \cite{macraild2022systematic} to filter out drugs without publically available assay data and FDA/TGA approval. Next, we use a clinical trials filter to divide our remaining 1029 drugs into two groups of 296 (undergoing clinical trials for COVID-19) and 733 (not under clinical trials for COVID-19) drugs. These 733 drugs are further passed through a set of 7 filters designed by Macraild et al. \cite{macraild2022systematic}. These 7 filters are as follows: 

\begin{enumerate}
    \item CC\textsubscript{50} or Selectivity Index
    \item COVID-19 IC\textsubscript{50}
    \item Cationic Amphiphilic Drugs (CAD) \& Pan-Assay Interference (PAINS)
    \item Route of Administration 
    \item Pregnancy 
    \item Black box warning 
    \item Indication
\end{enumerate}

\begin{figure}
    \begin{center}
    \includegraphics[scale=0.85]{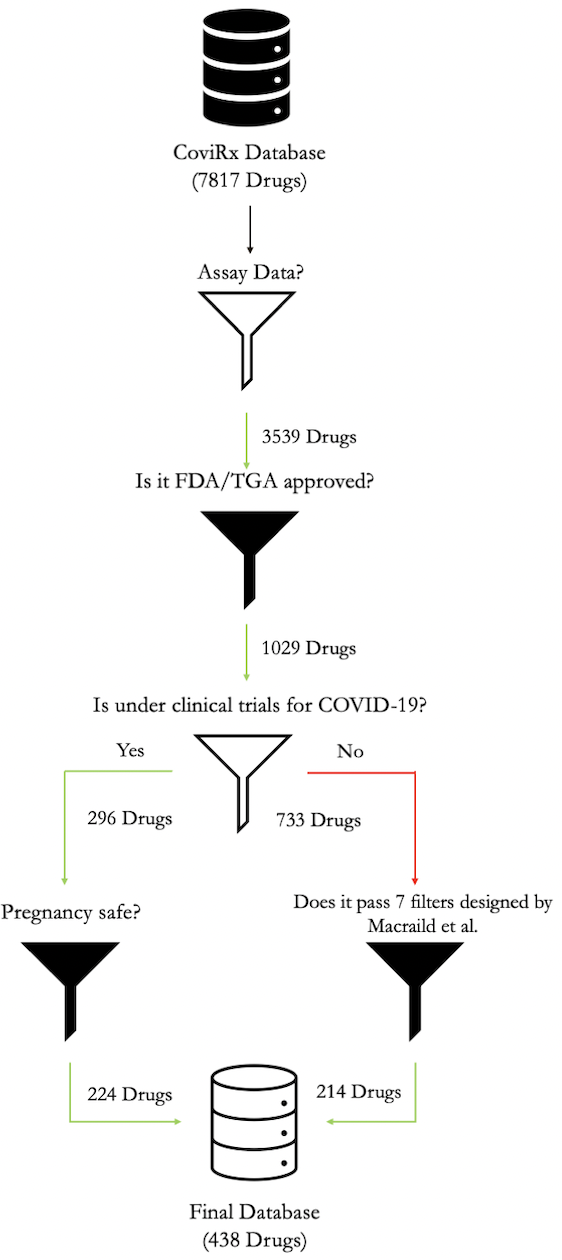}
    \end{center}
    \caption{Methodical selection of 438 drugs from 7817 drugs through filters designed by Macraild et al. \cite{macraild2022systematic}.}
    \label{fig:drug-selection}
\end{figure}

Through these filters, we remove cytotoxic drugs (CC\textsubscript{50} value less than 10µM) and pharmacologically inactive drugs such as supplements \& diagnostic tools. We also remove the drugs with poor IC\textsubscript{50} value (10x original indication or more) and unsafe for use during pregnancy (category D and X). Furthermore, we remove compound classes that cause PAINS, drugs with serious side effects, and CADs.  Figure \ref{fig:drug-selection} describes the process of selecting 438 drugs in detail. The selected drugs have over 100 features, and processing these features for the machine learning framework was challenging due to different feature types and interrelation. We present the details of the data processing and encoding in the following section. 

\subsection{Data processing and encoding} 

CoviRx comprises over 100 different features for the selected 438 drugs. These features were reduced to 68 with the help of domain experts to improve the quality of data for the machine-learning model. Furthermore, we remove information corresponding to the 8 filters designed by Macraild et al. \cite{macraild2022systematic}, (such as clinical trial information and drug's pregnancy category)  to reduce redundancy in downselection steps. We separate the selected features  into 4 textual features and 64 numerical features. Hence, we present the different textual features and their importance in drug repurposing research as follows:

\begin{enumerate}

  \item \textit{Drug-Mechanism of Action (MoA)}: The MoA is a specific biochemical interaction through which the drug produces an effect on the body \cite{moaInfo}. It can be useful in understanding how a drug will affect the body and identifying safe dosage levels for the drug.
  
  \item  \textit{Drug-Pathway}: A pathway in biology can be defined as a sequence of interactions between different biochemical molecules that gives rise to a certain product or changes the state of the cell \cite{fuzi2021path4drug}. By studying the different pathways through which a drug can act, one can identify new therapeutic uses for the drug by understanding the MoA for the drug and its metabolism \cite{zeng2015drug}. 
  
  \item  \textit{Drug-Indication}: The indication refers to a valid reason to use a particular drug.  This feature can help us analyze if a certain category of drugs, such as  antiviral drugs, is more likely to be effective against COVID-19. 
  
  \item  \textit{Drug-Target}: In biology, a target refers to anything within an organism to which another entity binds, thereby changing its behaviour/function \cite{zuanh}. Gene/protein targets of a drug can help identify if the drug will target the right entities to combat the effect of the virus on the host. 
  
\end{enumerate}

These four features represent network information and give insights into the interaction of the drug with various biological entities. Since traditional machine learning techniques such as label encoding and one-hot-vector encoding  cannot efficiently capture the network information and their interrelation, we have separated these 4 features from the rest of the numerical features. These have been encoded as 4 separate relationship matrices (0-1 matrix); where 0/1 represents the absence/presence of that feature. Table \ref{table:feature-info} presents the matrices and their dimensions. 

\begin{table}
\centering
\begin{tabular}{|p{0.75cm}|p{4cm}|p{2cm}|} 
 \hline
 S. No & Relationship Matrix & Dimension \\ [0.5ex] 
 \hline
  1 & Drug-MoA & 371 x 177\\ 
  2 & Drug-Pathway & 323 x 134\\
  3 & Drug-Indication & 435 x 180\\
  4 & Drug-Target & 328 x 626 \\ [1ex] 
 \hline
\end{tabular}
\caption{Different relationship matrices and their dimensions.}
\label{table:feature-info}
\end{table}

Additionally, we include both pathway and MoA information, even though these two are interdependent. The pathway can be considered a subset of MoA information because the signalling pathway followed by a drug directly affects its MoA. However, pathway information is present for some drugs, but MoA categorization is missing, and vice-versa. Hence, we decided to include both features.

The remaining 64 features contain information on the drug's physical and chemical properties, pharmacokinetics, and COVID-19 assay data points (IC\textsubscript{50}, CC\textsubscript{50}, selectivity index, \% viral inhibition, EC\textsubscript{50}, AUC, potency, efficacy, toxicity, and cytotoxicity from the different publically available assays as described  \cite{macraild2022systematic}). These features have been used as node features in the graph modelling step, details of which have been mentioned in the following sections. 

\subsection{Machine learning}
\subsubsection{Clustering methods}

Clustering is an unsupervised machine learning  method that deals with the segregation (grouping) of data based on the similarity  of data instances using given distance measures \cite{xu2015comprehensive}. The basic idea for clustering is that the data points within the same cluster must be as similar as possible, and the data points from different clusters must be as dissimilar as possible \cite{jain1988algorithms}. There are a number of different clustering algorithms, each with its own set of advantages and disadvantages. In machine learning and optimisation, it is well known that no single method is  suitable for all datasets, and often the nature of the problem determines their applicability \cite{gomez2016empirical,adam2019no}. Hence, we compare  the performance of three prominent clustering algorithms, namely \textit{k-means clustering} \cite{steinley2006k},  \textit{agglomerative hierarchical clustering} \cite{murtagh1983survey} and  \textit{spectral clustering} \cite{donath1973lower}.


K-means clustering  divides a data set into $k$ non-overlapping clusters, i.e., each data point belongs to only one cluster. It is one of the oldest and most prominent clustering techniques used in a wide range of problems \cite{forgy1965cluster}. The algorithm takes a dataset and a variable $k$ and returns a list of cluster labels for each data point (instance) in the dataset. The variable $k$ refers to the number of centroids in the data set where each centroid represents the cluster centre \cite{KMCInfo}.  In the first step of the algorithm, also known as the \textit{extension step}, we assign each data point  to its nearest centroid. In the next step, also known as  the \textit{maximization step}, we compute the mean for all points in a cluster and update the centroid. The algorithm stops once the centroids stabilize according to an error precision or when the maximum number of iterations has been reached \cite{KMCInfo2}. We use the \textit{residual sum of squares} (RSS) error to evaluate the quality of clusters with the goal of reducing it over time (iterations). We define the RSS  as the sum of the squared Euclidean distance of each data point from its centroid \cite{RSS}. The \textit{elbow method} is a heuristic used in cluster analysis for computing the optimum number of clusters. The main idea is to choose a point such that the diminishing returns are not worth the added cost, i.e., adding new clusters should not lead to better data modelling. In the past, k-means clustering has been used for a variety of problems in the biomedical domain, such as clustering patient disease data \cite{silitonga2017clustering}, identifying similarities among different Alzheimer's disease patients  \cite{alashwal2019application} and pattern discovery in healthcare data \cite{haraty2015enhanced}.
 
Agglomerative hierarchical clustering, also known as agglomerative nesting (AGNES), is a greedy algorithm that constructs a hierarchical relationship among the data points to cluster the data \cite{johnson1967hierarchical}. The   hierarchical structure is typically represented using a  tree structure known as a dendrogram which is constructed by applying a sequence of irreversible steps \cite{murtagh2012algorithms}. In the beginning, a cluster consists of a single data point. In the next step, most neighboring two clusters are merged to form a new cluster, and this process continues until only one cluster remains \cite{xu2015comprehensive}. AGNES can be divided into two groups, linkage methods and methods that specify cluster centres \cite{murtagh2012algorithms}. AGNES also visually represents the hierarchical relationship among clusters through dendrograms. The dendrograms can be further analyzed to study the relationship between data points and to select the optimal value of $k$ for clustering. In the past, hierarchical clustering has been used for a variety of problems in the biomedical domain, such as evaluating the level of preparedness for COVID-19 in 180 studies \cite{sadeghi2021using}, analyzing  co/multi-morbidities \cite{singh2018agglomerative}, and predicting 1-year mortality after hemodialysis \cite{komaru2020hierarchical}.

Spectral clustering is implemented as a graph partitioning problem where graph vertices represent data points that make no assumptions about the shape or form of the clusters \cite{SCInfo}. It  performs dimensionality reduction on the eigenvalues of the similarity matrix of the data before clustering. The goal   is to identify similar vertices in a graph based on connections between them \cite{von2007tutorial}. The graph is partitioned so that high-weight edges connect groups with similar features, and the dissimilar data points are connected by low-weight edges \cite{xu2015comprehensive}. The algorithm begins  by taking as input the number of clusters $k$ and projecting input data points with pairwise similarity into a graph (similarity matrix). Hence, the objective is to model the local neighbourhood relationship between data points while constructing the similarity matrix. There are three popular constructions; namely, epsilone-neighbourhood graph, k-nearest neighbour graph, and fully connected graph for the transformation \cite{von2007tutorial}. In the next step, the algorithm calculates  the graph Laplacian to a similarity matrix. Next, it solves the eigenvalue problem and selects $k$ eigenvectors corresponding to $k$ lowest/highest eigenvalues to define a k-dimensional subspace   using k-means clustering \cite{SCInfo}. In the past, spectral clustering has been used for a variety of problems in the biomedical domain, such as clustering single and multi-omics data \cite{john2020spectrum}, analyzing medical records \cite{yalccin2015analysis} and functional magnetic resonance imaging (fMRI) segmentation \cite{kuo2014spectral}. 

It is essential to use appropriate metrics to evaluate different clustering methods. It is difficult to validate the accuracy and quality of clusters formed by the clustering algorithms. The silhouette score can be used in cases where true labels are absent. The \textit{silhouette score}  measures the validity of clustering techniques by  quantifying  how similar an object is to its own cluster (cohesion) when compared to other clusters (separation) \cite{ROUSSEEUW198753}. A silhouette score of 0 indicates that the clusters are indifferent, and a value of -1 indicates that the clusters have not been formed properly.  A value of 1 means that the clusters are clearly distinguishable. Hence, we can evaluate the quality of clusters and, in some cases, can also  compute the optimal number of clusters. The silhouette score  is given by  $(x-y)/max(x,y)$; where $x$ and $y$ are   the  average intra-cluster distances. The silhouette score has been widely used in bioinformatics and medical applications, such as comparing clustering algorithms for healthcare datasets \cite{ogbuabor2018clustering} and identifying groups in gene expression data for analyzing differential expression results \cite{zhao2018silhouette}.

\subsubsection{Graph Neural Networks}
Graph neural networks are machine learning models  for processing data  in non-Euclidean data that is  represented as a graph \cite{4700287}. Graph neural networks have been prominent  in a variety of real-world problems with  graph (network) datasets such as  social media \cite{bian2020rumor}, community networks \cite{liu2020deep}, and citation networks \cite{frank2019evolution}. Given the increase in the availability of biological network data, graph neural networks have been used extensively in bioinformatics to tackle a wide range of problems such as disease prediction \cite{zhang2021graph}, polypharmacy side-effects prediction \cite{zitnik2018modeling}, and drug-disease treatment \cite{wu2017predicting}. 

Graph neural networks can be used to generate node embeddings (feature engineering) using autoencoders  which be used for  machine learning tasks \cite{wu2020comprehensive}. In addition, their application in prediction and classification tasks for graph-based data is also prominent \cite{pires2014mcsm,pham2022graph,das2022graph}. Node embeddings are an efficient way to generate features based on the interaction between different data points in a data set that eliminates the need for domain knowledge for feature engineering. In a graph-based data representation, every node represents a data point that may have a set of associated features. The node embedding represents its relationship with other records in the data. Thus, node embeddings capture the interaction of a node with its neighbouring nodes by learning from the input feature of the node itself and its neighbours and encoding this information into a vector in a latent space \cite{hamilton2017representation}. Different graph-based machine learning models can be used to learn these embeddings, such as traditional graph neural networks \cite{4700287}, graph convolution networks \cite{kipf2016semi}, graph attention networks \cite{velivckovic2017graph}, and variational graph autoencoders \cite{kipf2016variational}. 

\subsubsection{Variational graph autoencoders}  

The canonical autoencoder is a feedforward neural network that has two components that include  an \textit{encoder} and a \textit{decoder} \cite{baldi2012autoencoders}. The encoder converts an input data point $\bf x$ into a lower-dimensional embedding $ \bf y$, and the decoder remodels this embedding $\bf y$ into $\hat{\bf x}$ which is similar to $\bf x$. The low dimensional embeddings can be used directly for various downstream machine learning tasks such as classification. These low-dimensional embeddings are useful as these would require less storage and computational resources. Variational autoencoders (VAE)\cite{kingma2019introduction} offer the added advantage of generating new data from the original data point over traditional autoencoders. A variational autoencoder is a probabilistic generative model that embeds  the input to distribution and then a random sample $\bf y$ is taken from this distribution instead of generating it directly from an encoder \cite{kingma2013auto}. Hence, the encoder can produce multiple samples that come from the same distribution, enabling model uncertainty quantification in the low-dimensional embedding.  

Variational graph autoencoders (VGAE) feature a graph convolution network as its encoder, and a simple inner product decoder \cite{kipf2016variational}. The loss function for VGAE has two parts where the first part features the reconstruction loss between input (adjacency matrix $\bf A$ and feature matrix $\bf x$) and reconstructed lower-dimension embedding $\bf y$. The second part is the KL-divergence \cite{KLInfo} between $q(\bf  y|\bf x,A)$  and  $p(\bf y)$, where $p(\bf y) = N(0,1)$ that measures how close $q(\bf y|\bf x,A)$ is to $p(\bf y)$ \cite{graph2Info,kipf2016variational}. This architecture has been shown to perform extremely well for unsupervised learning, such as link prediction tasks. In addition, it can also be used for generating high-quality node embeddings \cite{kipf2016variational} and used for medical applications \cite{ding2021variational}.

In our proposed framework, we implement feature engineering using  graph data  using variational graph autoencoders. The feature embedding is  later used as input feature vectors for  clustering the data.  In our framework, we will compare VGAE with graph autoencoders (GAE) to select the better-performing model for generating the node embeddings for each drug (node).
 
\subsection{Framework}

\begin{figure*}
\begin{center}
    \includegraphics[width=0.60\textwidth]{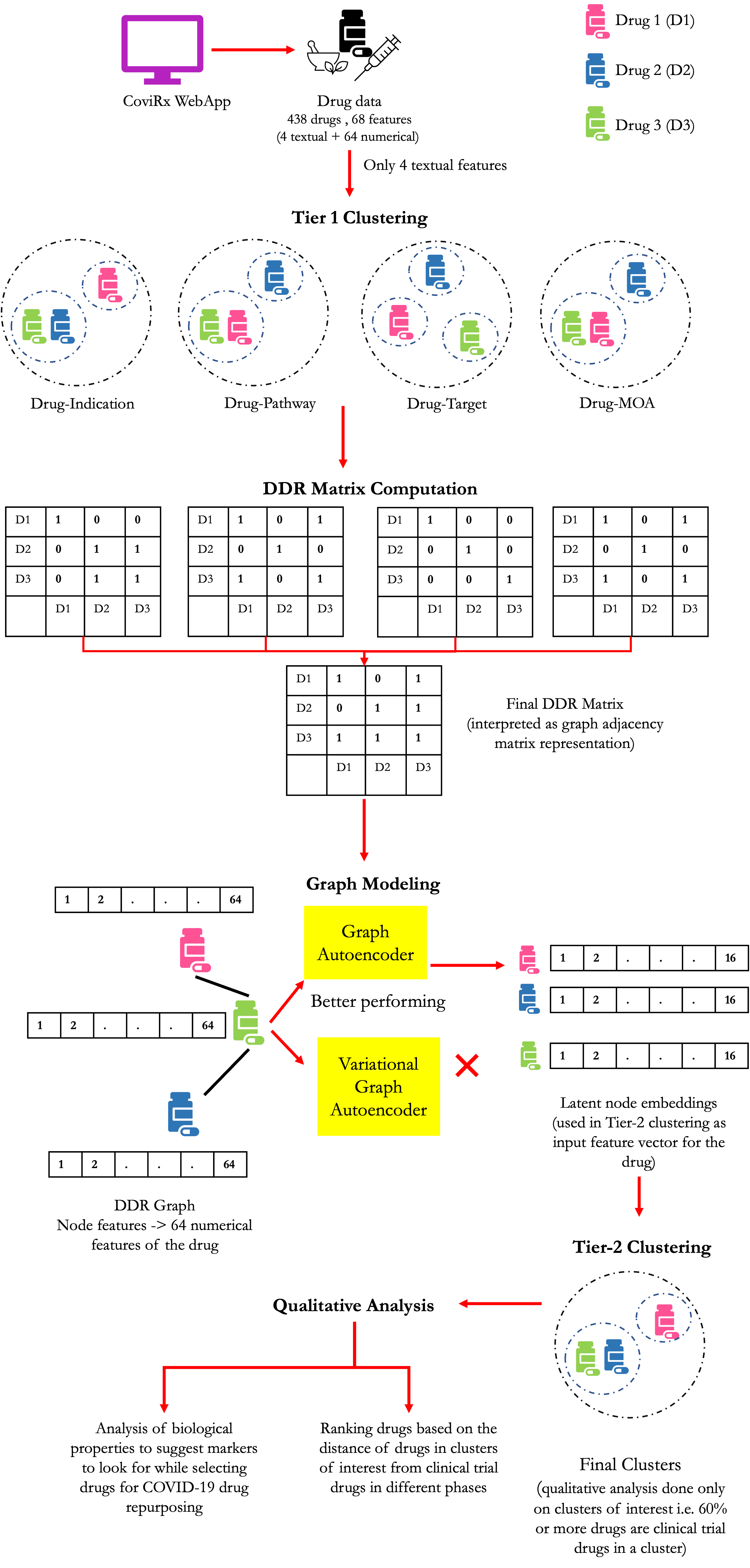}
\end{center}
    \caption{Proposed unsupervised machine learning framework for heterogeneous clustering of drugs for COVID-19. We applied this framework to drugs downselected in Figure \ref{fig:drug-selection}.}
    \label{fig:framework}
\end{figure*}

The recommendation of  drug candidates from the CoviRx database for COVID-19 drug repurposing can be viewed as a multi-feature type clustering problem as drug features are heterogeneous. In this study, we propose an unsupervised clustering framework that groups drugs with similar therapeutic properties and identifies new therapeutic properties of a given drug depending upon the therapeutic properties of its neighbours (related drugs). To effectively integrate heterogeneous data, we need  a two-tier clustering framework.

Figure \ref{fig:framework} presents an overview of the framework  that considers a collection of heterogeneous features in the data. We adopt a 2-tier clustering approach inspired by Hameed et al. \cite{hameed2018two}, who used a similar two-tier framework for clustering the drugs based on their Anatomical Therapeutic Chemical (ATC) classification for drug repurposing. They used this framework to find out if a drug can belong to more than one ATC class and thus have more than one therapeutic use. We implement Tier-1 clustering  based on homogeneous features, and in Tier-2, we combine the output (results) from Tier-1 to form a single drug-drug relationship matrix. This matrix is then modelled as a graph with remaining drug features as input node features to generate node embeddings in an unsupervised fashion and cluster the drugs. Our  2-Tier clustering framework employs graph autoencoders so that the final drug clusters are formed based on the interactions among all 68 drug features and not just on the absolute values of these features.

In Step 1 (Tier-1 Clustering), we cluster the drugs according to their homogeneous biological properties: pathway, target, MoA, and indication. We merge the  clusters obtained from Step 1 to form a drug-drug-relationship (DDR) matrix  in Step 2 (DDR Matrix Creation). Then in Step 3 (Graph Modelling), we use graph autoencoders to create a final feature vector for each drug. In Step 4 (Tier-2 Clustering), the feature vectors from Step 3 are used to get the final clusters of drugs which are then evaluated qualitatively. 

\subsubsection{Tier-1 Clustering}

We consider each row in a relationship matrix as an input feature vector while applying the different clustering algorithms. We evaluate the quality of the clusters using a silhouette score where the clusters with the highest silhouette scores are considered for further evaluation. The selected clusters are then used to generate 4 different DDR matrices of varying sizes for four different textual features, i.e.,  size of 371 x 371 (MoA), 323 x 323 (pathway), 435 x 435 (indication), and 328 x 328 (target). Table \ref{table:feature-info} provides information about  dimensions of original matrices \textit{\textbf{N x M}}, where \textbf{N} refers to the number of drugs for which the information is present and \textbf{M} refers to the number of unique feature values in the data. These DDR matrices are 0-1 matrices, where the entry for a cell [i,j] is 1 if drug\textsubscript{i} and drug\textsubscript{j} belong to the same cluster, and 0 otherwise. 

We extended the dimensions of the DDR1 matrices  to 438 x 438 by adding the missing drugs, where each missing drug is considered a separate cluster (the number of missing information per feature has been described in Table \ref{table:feature-info}). The dimension extension step was not done before clustering to avoid grouping drugs with missing data in the same cluster.

\subsubsection{Drug-Drug-Relationship Matrix}

We note that we have four different categories of matrices (MoA, pathway, indication, and target) and need to combine them to ensure that we fuse the data in our framework. 
We combine the four DDR matrices  using the logical OR function to get the final DDR matrix as explained by the DDR matrix computation step in Figure \ref{fig:framework}. We implement this step to combine the features from different homogeneous spaces into one common heterogeneous space, as this would maximize information gain. We utilize  logical OR  over logical AND as drugs that are not clustered together in all four feature spaces will not be represented in the final DDR matrix, thus leading to loss of information. Hence, the final DDR matrix  models the information about the interaction of the drug with different biological entities as drug-drug connections. 

\subsubsection{Graph Modeling}

We next consider the final DDR matrix  as an adjacency matrix representation for the DDR graph. In this graph, each node (drug) has a set of features associated with it, which represent the  64 numerical features of the drug that have been previously separated (data processing step). We generate low-dimension latent vector embeddings for all our drug nodes using graph autoencoders (GAE) in this step. 
The goal here is  to generate embeddings by combining the feature information of a node with its neighboring nodes. The final node embeddings now encode all the different drug features, such as PK/PD, assay information, and biological, physical, and chemical properties in a low-dimension vector which can be used for clustering.

\subsubsection{Tier-2 Clustering}

In Tier-2 clustering,  we use the node embeddings obtained from the graph modeling step as the input feature vector for each drug. We then apply the clustering algorithms that include k-means,   spectral, and hierarchical clustering  to generate the final drug clusters. We evaluate the quality of clusters using the silhouette score, and clusters with the highest silhouette score will be considered for further evaluation. Once we have selected the clusters, we define our clusters of interest. These clusters feature  the drugs that passed the \textit{clinical trial filter} and the drugs that failed the filter. The clusters contain the drugs that are very close to drugs that have undergone clinical trials for COVID-19 and thus can be explored as a starting point for COVID-19 drug repurposing experiments. We evaluate these clusters   qualitatively by reviewing the available literature.
 
\subsection{Implementation}

In our framework, we  implement the clustering algorithms using scikit-learn \cite{scikit-learn} library in Python. We use another Python library, TensorFlow \cite{tensorflow2015-whitepaper}, to tweak the original implementation of VGAE \cite{kipf2016variational} to suit our requirements. We use libraries such as  Graphviz\footnote{\url{https://graphviz.org/}} \cite{gansner2000open} and networkx\footnote{\url{https://networkx.org/}} \cite{hagberg2008exploring} to make graph visualization. We develop  the entire pipeline  in Python 3.8 and use Google Colab\footnote{https://colab.research.google.com/} to run our experiments where only the graph modelling step makes use of Colab graphics processing units (GPU). 

\section{Results} 

\subsection{Tier-1 Clustering}

\begin{table}
\small
    \begin{subtable}{0.45\textwidth}
        \centering
            \begin{tabular}{|p{1.5cm}|p{1.75cm}|p{1cm}|p{1cm}|p{1.25cm}|} 
             \hline
             Feature & Number of Clusters & Elbow & Inertia & Silhouette Score \\ [0.5ex]
             \hline
             MoA & $k=50$ & -- & 166.46 & 0.44\\ 
             & $k=100$ & -- & 86.36 & 0.53 \\
             \rowcolor{LightCyan}
             & $k=150$ & Yes & 19.96 & 0.56 \\
             & $k=200$ & -- & 2.96  & 0.48 \\ [1ex] 
             \hline
             Pathway & $k=50$ & -- & 158.99 & 0.39\\ 
             & $k=100$ & -- & 55.63 & 0.48 \\
             \rowcolor{LightCyan}
             & $k=160$ & Yes & 5.95 & 0.52 \\
             & $k=200$ & -- & 0.65  & 0.44 \\ [1ex] 
             \hline
             Target & $k=50$ & -- & 497.52 & 0.08\\ 
             & $k=100$ & -- & 265.99 & 0.17 \\
             \rowcolor{LightCyan}
             & $k=150$ & -- & 139.75 & 0.24 \\
             & $k=200$ & -- & 61.46  & 0.26 \\ [1ex] 
             \hline
             Indication & $k=50$ & -- & 203.30 & 0.44\\ 
             & $k=100$ & -- & 99.69 & 0.51 \\
             \rowcolor{LightCyan}
             & $k=180$ & Yes & 16.34 & 0.54 \\
             & $k=200$ & -- & 6.80  & 0.52 \\ [1ex] 
             \hline
            \end{tabular}
            \caption{K-Means Clustering Results}
            \label{table:kmeans-t1}
    \end{subtable}
    \hfill
    \begin{subtable}{0.45\textwidth}
        \centering
            \begin{tabular}{|p{1.5cm}|p{2cm}|p{2cm}|p{1.25cm}|} 
             \hline
             Feature & Number of nearest neighbors & Number of Clusters & Silhouette Score \\ [0.5ex]
             \hline
             MoA & $n=7$ & $k=56$ & 0.63\\ 
             \rowcolor{LightCyan}
             & & $k=81$ & 0.71 \\ 
             & & $k=253$ & 0.45 \\ [1ex]
             \hline
             Pathway & $n=5$ & $k=5$ & 0.18\\  
             & & $k=25$ & 0.33 \\ 
             \rowcolor{LightCyan}
             & & $k=70$ & 0.58 \\ [1ex]
             \hline
             \rowcolor{LightCyan}
             Target & $n=10$ & $k=7$ & 0.22\\ 
             & & $k=48$ & 0.13 \\
             & & $k=76$ & 0.17 \\ [1ex]
             \hline
             Indication & $n=7$ & $k=16$ & 0.33\\  
             \rowcolor{LightCyan}
             & & $k=194$ & 0.64 \\ 
             & & $k=199$ & 0.64 \\ [1ex]
             \hline
            \end{tabular}
            \caption{Spectral Clustering Results}
            \label{table:spectral-t1}
     \end{subtable}
     \hfill
    \begin{subtable}{0.45\textwidth}
        \centering
            \begin{tabular}{|p{2cm}|p{2.75cm}|p{2.5cm}|} 
             \hline
             Feature & Number of Clusters & Silhouette Score \\ [0.5ex]
             \hline
             \rowcolor{LightCyan}
             MoA & $k=80$ & 0.72\\ 
             & $k=140$ & 0.71\\
             & $k=200$ & 0.35\\
             \hline
             Pathway & $k=80$ & 0.63\\  
             \rowcolor{LightCyan}
             & $k=140$ & 0.72\\
             & $k=200$ & 0.32\\
             \hline
             Target & $k=80$ & 0.18\\ 
             \rowcolor{LightCyan}
             & $k=160$ & 0.34\\
             & $k=200$ & 0.32\\
             \hline
             Indication & $k=80$ & 0.65\\ 
             \rowcolor{LightCyan}
             & $k=180$ & 0.73\\
             & $k=200$ & 0.65\\
             \hline
            \end{tabular}
            \caption{Hierarchical Clustering Results}
            \label{table:hierarchical-t1}
     \end{subtable}
     \caption{We report the silhouette score for different values of $k$ for Tier-1 Clustering and  highlight the best-performing combinations.}
     \label{tab:T1}
\end{table}

\subsubsection{K-means clustering}

We use the \textit{elbow method} and silhouette score in combination to determine  the appropriate value of $k$ and evaluate the quality of clusters for k-means clustering. We observe that, on average, the minimum value of inertia and the maximum value of silhouette score   for $k=150$. Table \ref{table:kmeans-t1} presents inertia and silhouette scores for different values of $k$, which indicates that k-means clustering requires large values of $k$ to form meaningful clusters from the data. 

\subsubsection{Spectral Clustering}

In the case of spectral clustering, we use the \textit{eigengap} heuristics approach based on perturbation theory and spectral graph theory  to compute the top-performing number of clusters. According to this approach, the value of $k$ should maximize the difference between consecutive eigenvalues (eigengap) \cite{von2007tutorial}. It indicates that the spectral clustering algorithm performs better if the eigenvectors are closer to each other, which happens when the eigengap is large. This algorithm takes as input the affinity matrix of the input data, which has been calculated by applying local scaling based on $k$ nearest neighbours \cite{zelnik2004self,spectralInfo}. Furthermore, we also use the silhouette score   to evaluate the quality of the clusters. Table \ref{table:spectral-t1} presents the top 3 optimal numbers of clusters with the number of nearest neighbours and their corresponding silhouette scores.

\subsubsection{Hierarchical Clustering}

We apply hierarchical clustering to further compare the performance with the previous clustering methods. Due to the large number of drug samples in the dataset, the dendrograms obtained from hierarchical clustering  were difficult to analyze; hence, they could not be used for calculating the number of optimal clusters. Therefore, we evaluated  the silhouette scores for cluster sizes in the range [50,200] with a jump size of 20 and selected the clusters with the maximum score. Table \ref{table:hierarchical-t1} presents the silhouette score for the optimal number of clusters. We observe that  except for MoA, the optimal cluster size is close to 150, which was also observed for k-means clustering.

\subsection{Graph Modeling}

\begin{figure*}
\begin{center}
    \includegraphics[width=\textwidth]{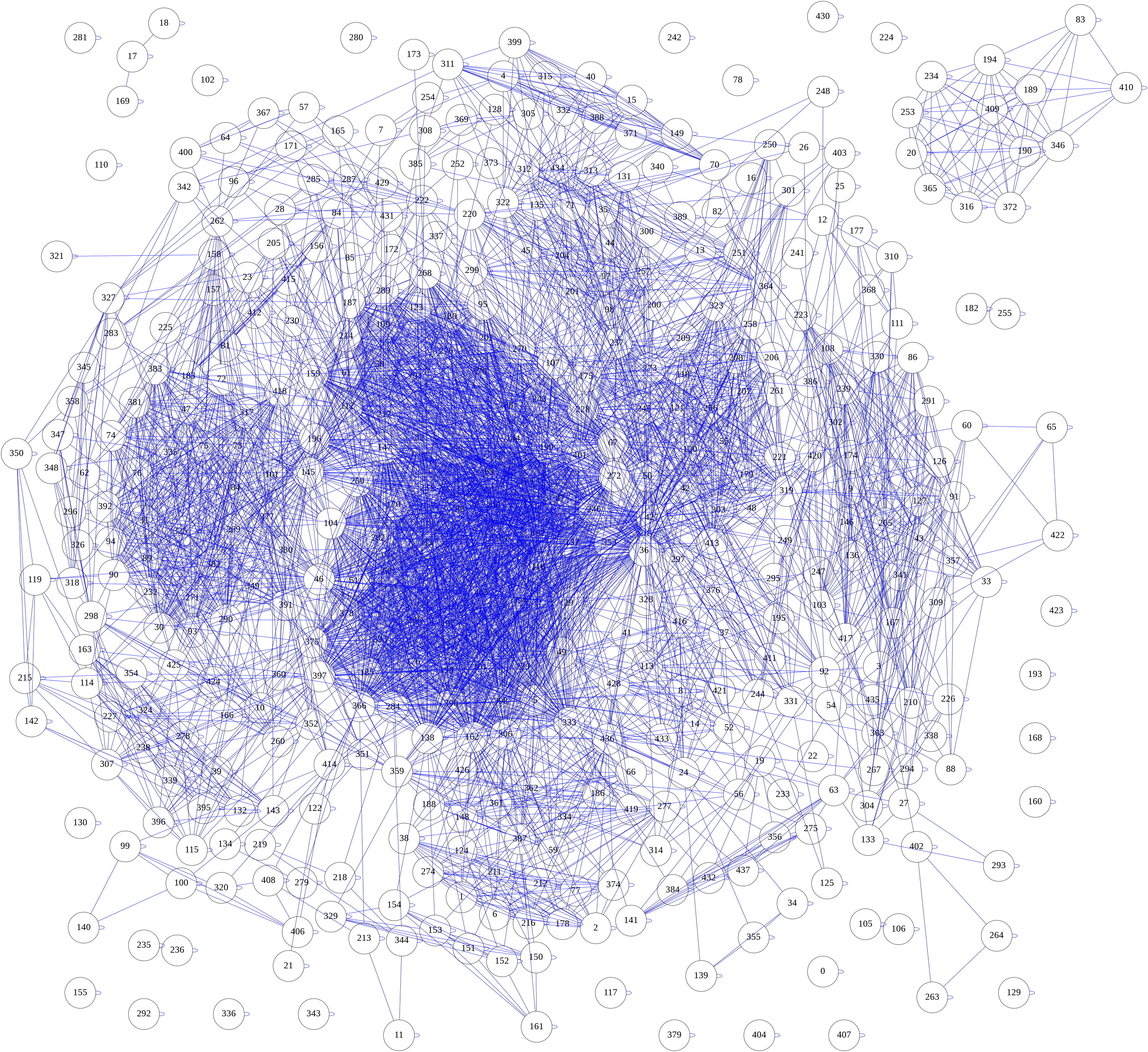}
\end{center}
    \caption{Graph derived from final DDR matrix that was formed by combining  Drug-Indication DDR1 matrix, Drug-Pathway DDR1 matrix, Drug-MoA DDR1 matrix, and Drug-Target DDR1 matrix through the logical-OR operator} 
    \label{fig:DDR-final}
\end{figure*}

We observe that the clusters formed by hierarchical clustering have maximum silhouette scores; thus, these are taken forward for further analysis. 
We combined the four DDR matrices obtained after Tier-1 clustering  to form a final DDR matrix using the logical OR function. We thus obtained the final DDR matrix that has a sparsity of 95\%. Figure \ref{fig:DDR-final} presents the graph formed by the final DDR matrix (adjacency matrix representation), which shows a big group of drugs with similar biological properties. We use both GAE and VGAE to input this graph data, which contains numerical features as node feature vectors. We ran trail experiments and found that a hidden layer dimension of 128, an embedding size of 16, and a learning rate of 0.01 work best for our model (low loss value).   We also found  that GAE outperformed VGAE for this problem with a loss of 0.31 when trained for 500 epochs. The latent embeddings thus obtained were used for the next layer of clustering. These latent embeddings now encode information corresponding to the drug's textual and numerical features. The respective clustering algorithms treat the latent vector as an input feature vector for each drug. 

\subsection{Tier-2 Clustering}

Similar to Tier-1 Clustering, we use the elbow method  for k-means clustering and the eigengap heuristics approach  for spectral clustering to calculate the optimal number of clusters. However, in this tier, we observed the elbow for k-means is obtained at a lower value of $k$ (8-15 clusters). 

Unlike Tier-1 Clustering, the dendrogram we obtain after applying hierarchical clustering is much easier to interpret and we find that the optimal value of clusters lies in the range of [10,20]. Furthermore, we use  the silhouette score to evaluate the quality of the clusters  for the optimal value of clusters, as shown in Table \ref{table:t2}.

\begin{table}
\small
\centering
\begin{tabular}{|p{3.5cm}|p{2cm}|p{1.75cm}|} 
 \hline
 Method & Number of Clusters & Silhouette Score \\ [0.5ex]
 \hline
 K-means Clustering & $k=8$ & 0.35\\ 
 \rowcolor{LightCyan}
 & $k=10$ & 0.37 \\
 & $k=15$ & 0.34\\
 \hline
 Spectral Clustering & $k=32$ & 0.45\\  
 & $k=52$ & 0.45\\
 \rowcolor{LightCyan}
 & $k=63$ & 0.48\\
 \hline
 \rowcolor{LightCyan}
 Hierarchical Clustering & $k=10$ & 0.47 \\ 
 & $k=12$ & 0.46 \\
 & $k=18$ & 0.47 \\
 \hline
\end{tabular}
\caption{Comparison among different values of silhouette score for different values of $k$ for all clustering algorithms applied in Tier-2 Clustering. The best-performing combination has been highlighted.}
\label{table:t2}
\end{table}

\begin{figure}
\begin{center}
    \includegraphics[scale=0.4]{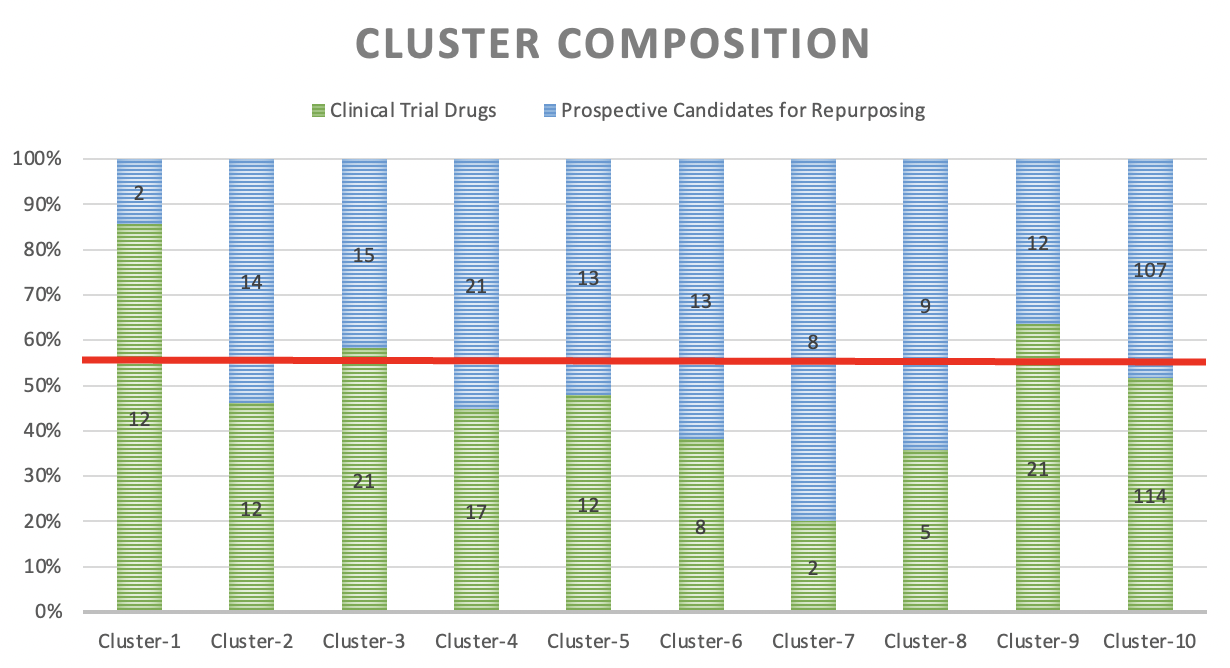}
\end{center}
    \caption{Composition of cluster set selected for qualitative analysis.}
    \label{fig:cluster-composition}
\end{figure}

We finally select the clusters obtained from hierarchical clustering with  $k=10$  for the final qualitative analysis. We prefer this over clusters from spectral clustering with a slightly higher silhouette score since the large value of $k$  makes qualitative analysis difficult. 

\subsection{Qualitative Analysis}

Finally, we apply qualitative analysis on the clusters obtained from the best clustering method in the previous section. Note that some of the drug samples in the clusters are already part of some stage of the clinical trial for COVID-19, i.e., Dexamethasone \cite{wu2022low} and Lopinavir that are in Stage 4 of clinical trial \cite{CTInfo}.
We define clusters of interest as the clusters with 55\% or more clinical trial drugs (left-hand side of Figure \ref{fig:drug-selection}) and the rest comprising prospective candidates for COVID-19 drug repurposing (right-hand side of Figure \ref{fig:drug-selection}). Figure \ref{fig:cluster-composition} presents the composition of different clusters comprising the final set of clusters selected for qualitative analysis. Since Cluster-1, Cluster-3, and Cluster-9 have 55\% or more clinical trial drugs,  we   qualitatively review them in order to recommend them for COVID-19 drug repurposing. 

\begin{figure}[htbp!]
\centering
\begin{subfigure}{0.475\textwidth}
    \includegraphics[width=\textwidth]{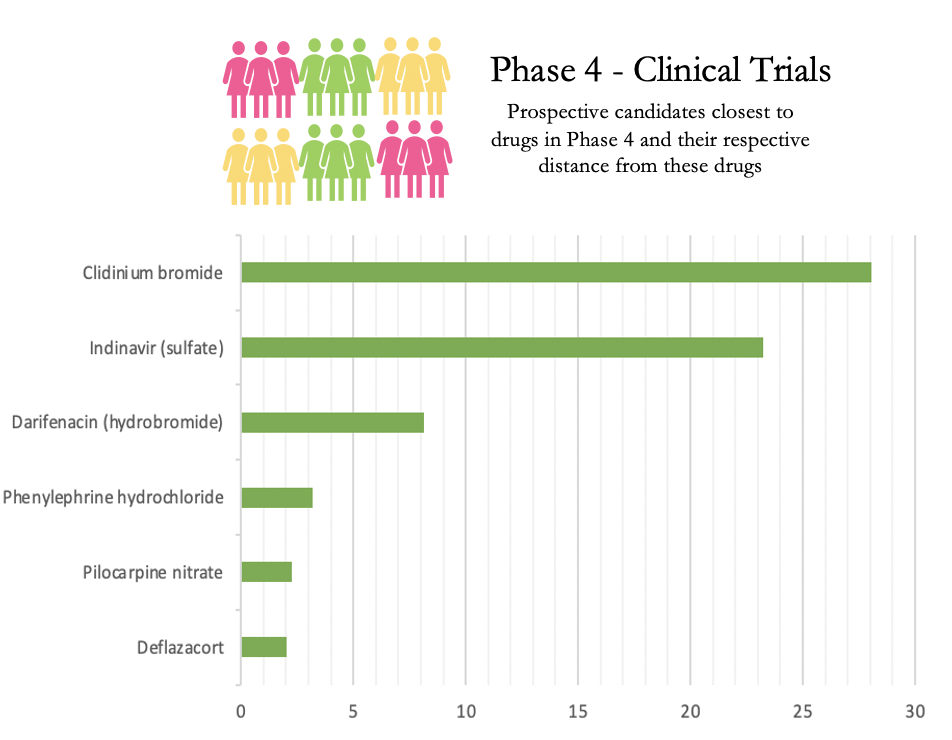}
    \caption{Phase-4 clinical trial drugs.}
    \label{fig:p4}
\end{subfigure}
\hfill
\begin{subfigure}{0.475\textwidth}
    \includegraphics[width=\textwidth]{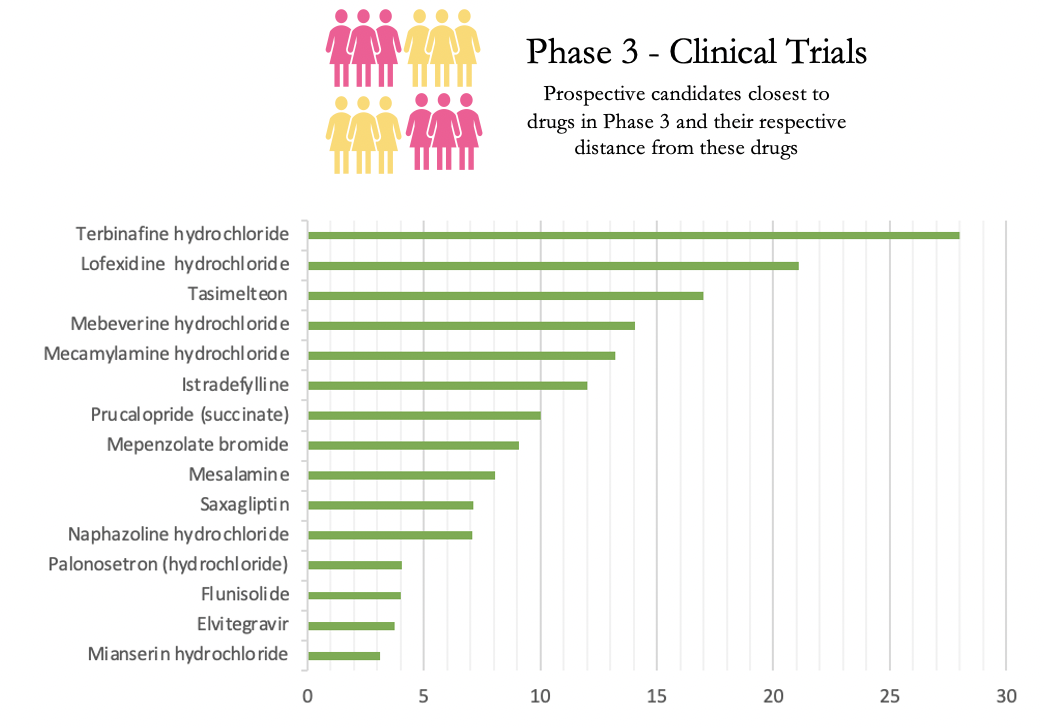}
    \caption{Phase-3 clinical trial drugs.}
    \label{fig:p3}
\end{subfigure}
\hfill
\vskip\baselineskip
\begin{subfigure}{0.475\textwidth}
    \includegraphics[width=\textwidth]{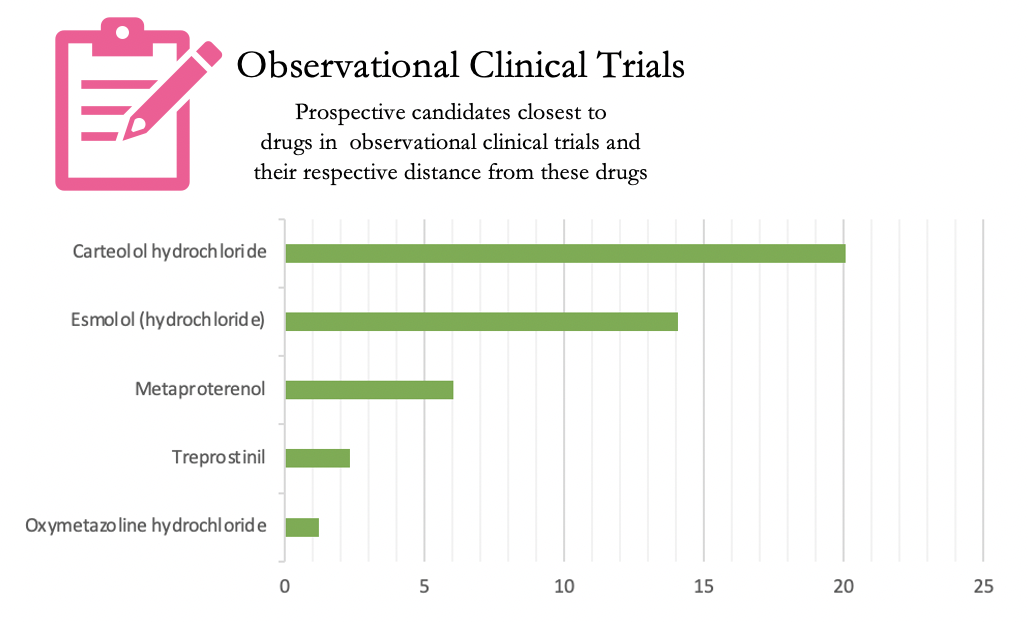}
    \caption{Observational clinical trial drugs.}
    \label{fig:obs}
\end{subfigure}
\caption{Euclidean distance of prospective candidates from their nearest clinical trial neighbors.}
\label{fig:cluster-analysis}
\end{figure}

We further categorize the prospective candidates present in clusters of interest into different groups depending upon their Euclidean distance from clinical trial drugs in different phases present in the same cluster (as themselves). Figure \ref{fig:p4} presents the prospective candidates from all three clusters of interest that are nearest to Phase-4 clinical trial drugs and their Euclidean distance from these drugs. Similarly, Figure \ref{fig:p3} presents the same data for Phase-3 clinical trial drugs, and Figure \ref{fig:obs} presents it for observational clinical trial drugs. We observe that only 1 drug is closest to Phase-1 clinical trial drugs, and find no drug   to be closest to Phase-2 clinical trial drugs. 

Next, we rank  the drugs according to their distance from clinical trial drugs in different phases and select the top 15 drugs. We assign the drugs closer to Phase-4 clinical trial drugs  a higher rank, followed by the ones closer to Phase-3, Phase-2, Phase-1, and finally, the observational phase. We rank drugs closer to Phase-4 clinical trial drugs   higher because clinical-4 trial drugs are in the later stages of study and thus have a higher chance of succeeding. Ranking within the same group (for example, closer to Phase 4) has been done on the basis of distance. The lesser the distance, the  higher the rank. Table \ref{table:drug-rank} lists the top 15 drugs along with their distance from drugs in different clinical trial phases to which they are closest.

\begin{figure}[htbp!]
\centering
\begin{subfigure}{0.475\textwidth}
    \includegraphics[width=\textwidth]{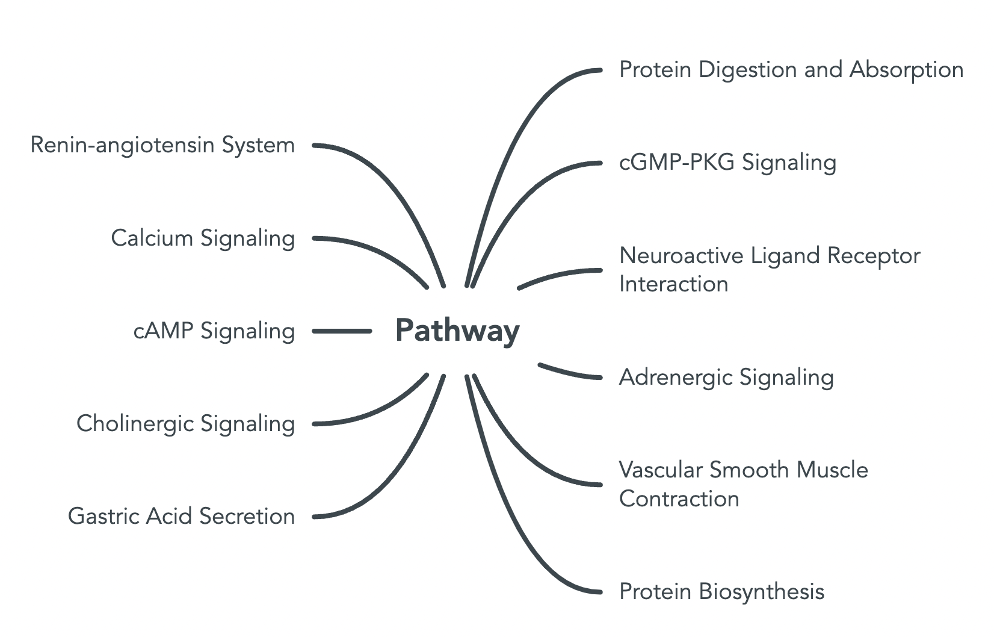}
    \caption{Pathway}
    \label{fig:path}
\end{subfigure}
\hfill
\begin{subfigure}{0.475\textwidth}
    \includegraphics[width=\textwidth]{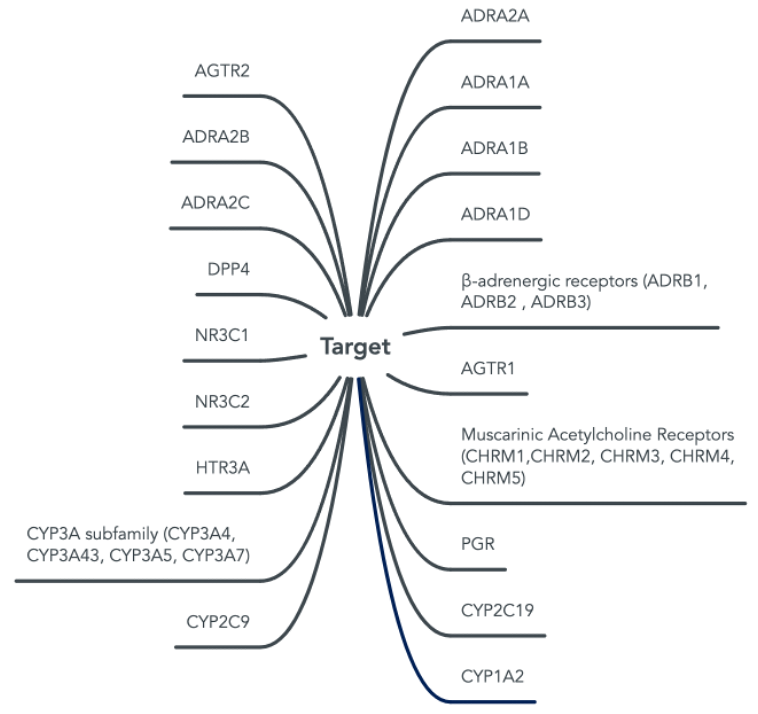}
    \caption{Target}
    \label{fig:target}
\end{subfigure}
\hfill
\vskip\baselineskip
\begin{subfigure}{0.475\textwidth}
    \includegraphics[width=\textwidth]{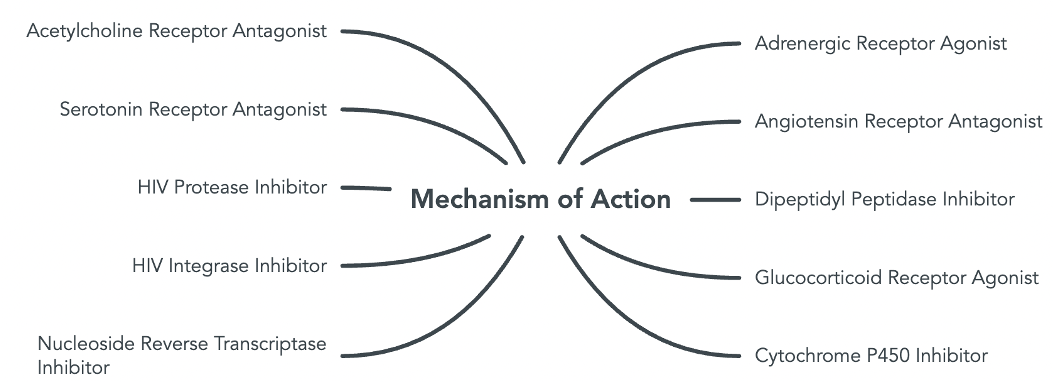}
    \caption{MOA}
    \label{fig:moa}
\end{subfigure}
\caption{Most commonly observed biological properties of drugs in clusters of interest.}
\label{fig:cluster-analysis}
\end{figure}

Finally, we analyze the different biological properties of these top 15 drugs and the remaining drugs in our clusters of interest. The most common biological properties (such as pathway, MoA, and target) can be treated as potential biological markers that one can look for when selecting drugs for COVID-19 drug repurposing in their own studies. Figure \ref{fig:path} presents the most commonly observed pathways in the clusters of interest. Similarly, Figure \ref{fig:target} presents the most commonly observed targets, and Figure \ref{fig:moa} presents the most commonly observed MoA. 

\begin{table}
\small
\centering
\begin{tabular}{|p{0.6cm}|p{4cm}|p{1.8cm}|p{1.0cm}|} 
 \hline
 Rank & Name of the Drug & Clinical trial group of nearest neighbor & Distance \\ [0.5ex]
 \hline
 1 & Deflazacort & Phase-4 & 2.00\\ 
 2 & Pilocarpine nitrate & Phase-4 & 2.26\\   
 3 & Phenylephrine hydrochloride & Phase-4 & 3.18\\ 
 4 & Darifenacin (hydrobromide) & Phase-4 & 8.12\\   
 5 & Indinavir (sulfate) & Phase-4 & 23.22\\ 
 6 & Clidinium bromide & Phase-4 & 28.05\\   
 7 & Mianserin hydrochloride & Phase-3 & 3.11\\ 
 8 & Elvitegravir & Phase-3 & 3.74\\   
 9 & Flunisolide & Phase-3 & 4.02\\ 
 10 & Palonosetron (hydrochloride) & Phase-3 & 4.03\\   
 11 & Naphazoline hydrochloride & Phase-3 & 7.08\\ 
 12 & Saxagliptin & Phase-3 & 7.10\\   
 13 & Mesalamine & Phase-3 & 8.06\\ 
 14 & Mepenzolate bromide & Phase-3 & 9.00\\   
 15 & Prucalopride (succinate) & Phase-3 & 10.02\\ 
 \hline
\end{tabular}
\caption{Top 15 drugs based on the distance from their nearest neighbours that are in different phases of clinical trials for COVID-19.}
\label{table:drug-rank}
\end{table}

\section{Discussion}
 
Even though vaccines have helped reduce the severity of the situation, COVID-19 has serious concerns as we do not have a safe and specific treatment.  Thus, there is a dire need to develop or repurpose drugs that can help with the treatment of COVID-19 and also help mitigate the long-term health effects of the virus. As noted earlier, drug discovery is a highly costly and complex process,   thus making drug repurposing an effective strategy for safe, effective, and affordable treatments. In this study, we made an effort  to identify prospective candidates for COVID-19 drug repurposing by using an unsupervised machine learning framework that uses graph representation learning for feature engineering. 

We used a  dataset comprising more than 7000 drugs   which were then down-selected to 485 drugs (details in Figure \ref{fig:drug-selection}) and used as input for the proposed framework. We developed a 2-Tier machine learning framework to combine the non-linear relationship among all drug properties and group the drugs into small high-quality clusters. In Tier-1, we observed that the value of \textit{k} was high for all the methods due to a larger number of unique values per feature (as mentioned in Table \ref{table:feature-info}). In Tier 2, the value of \textit{k} was significantly reduced as the algorithm now took into account the absolute feature values and the various type of interactions among these features. We found that out of the three clustering algorithms that were compared by the framework in Tier-1 and Tier-2 clustering, the hierarchical clustering algorithm outperformed spectral and k-means clustering by giving better quality clusters with a higher silhouette score. During the graph modelling step, GAE performed better than VGAE by giving high-quality drug node embeddings. 

Afterwards, we obtained  10 clusters by applying hierarchical clustering on the node embeddings obtained by GAE. Out of 10 clusters, 3  were defined as clusters of interest since 55\% of the drugs in these clusters were clinical trial drugs (details of cluster composition in Figure \ref{fig:cluster-composition}). When analyzed qualitatively, the drugs from these clusters resulted in the top-15 performing candidates, which were ranked on the basis of their Euclidean distance from drugs in different phases of clinical trials. The biological properties of the drugs in the clusters of interest were also analyzed to identify the most commonly occurring drug-pathway, drug-target, and drug-MOA properties (as listed in Figure \ref{fig:cluster-analysis}).  
 
We observed that the most common indications for drugs in clusters of interest were anti-HIV and anti-inflammatory. Previous studies have shown that anti-HIV drugs can be effective in lowering the risk of COVID-19 \cite{AntiHIVInfo}. These studies have further highlighted that anti-HIV integrase inhibition drugs can be prospective candidates for COVID-19 drug repurposing as these impede viral maturation and replication; however, additional experimental studies must be done to evaluate their efficacy \cite{salari2022anti}. HIV protease inhibitors on the other reduce viral load at high drug concentrations and thus might have low clinical importance \cite{mahdi2020analysis}. Nucleoside Reverse Transcriptase Inhibitor (for example, Remdesivir \cite{spinner2020effect}) prevent viral replication by blocking RNA-dependent RNA polymerase (RdRp) and thus can be effective in COVID-19 treatments \cite{frediansyah2021antivirals}. Indinavir (sulfate) and Elvitegravir, two anti-HIV drugs, have various in silico studies supporting their role in the treatment of COVID-19 infections. Indinavir (sulfate) \cite{pollak2022indinavir,indInfo} has been predicted to inhibit SARS-CoV-2 3CL-protease (3CLpro) and both of these have been predicted to impede RdRp activity  \cite{eltInfo,shimura2009elvitegravir}.

Studies have also shown that the soluble form of Angiotensin-Converting Enzyme 2 (ACE2) can inhibit SARS-CoV-2 infection. The Calcium signalling pathway mediates the release of the soluble catalytic ectodomain of ACE2 (ACE2 shedding). Thus, drugs enhancing ACE2 shedding can be used for treatment, provided appropriate experimental studies confirm their efficacy \cite{garcia2022calcium}. Similarly, glucocorticoids also improve the severity of COVID-19 infections by activating ACE2 \cite{xiang2020glucocorticoids}. The calcium signalling pathway also helps fight hypoxia and inflammation by relieving oxidative stress. Adrenergic receptor agonists have also been shown to be effective in lowering mortality due to hyperinflammation in COVID-19 as these can lower the inflammatory response and immune cells \cite{hamilton2021association}. Recent pathway enrichment analysis studies have also concluded that Neuroactive ligand-receptor interaction can be a pharmacologically important mechanism against COVID-19 \cite{oh2022network}. Dipeptidyl peptidase (DPP) inhibitors were thought to have no effect on the mortality rate among COVID-19 patients. However, recent studies have shown that DPP inhibitors can reduce mortality among COVID-19 patients. They can do so by several mechanisms, one of them being inflammatory response reduction \cite{aldukhi2021understanding}. 

The analysis of clusters of interest revealed HIV Integrase Inhibitors, Nucleoside Reverse Transcriptase Inhibitor, Adregenic receptor agonists, DPP inhibitors, Calcium signalling pathways, and Neuroactive ligand-receptor interaction as probable signalling pathways and mechanisms of action for some of the shortlisted drugs. This observation corroborates with reported studies on anti-viral mechanisms, and drug candidates that fall into this category of the mechanism of action/pathway may have the potential for anti-COVID-19 drug repurposing.
 
This study is a novel attempt to analyze data on CoviRx through machine learning models. 
\textcolor{black}{Our framework is novel since it employs clustering methods on graph-based data. The data features a combination of numerical and text data which was pre-processed and encoded into a graph structure where  autoencoders were used to create embedding that was further processed via three different clustering methods. This type of framework is unique to COVID-19 drug repurposing  which can enable better management of drug trials in future disease outbreaks and  pandemics. }
Previous drug repurposing studies on this data depended on expert knowledge for manual downselection \cite{macraild2022systematic}. Even though our study presents a novel framework, we have a  set of limitations. We note that CoviRx data was last updated on March 2022. In a follow-up study, the dataset can be updated with the help of domain experts before applying down-selection steps by doing an extensive literature review. Furthermore, the analysis can be enriched by including information about COVID-19 and its mechanism of causing infection within the framework itself.

\section{Conclusion}

This study developed a novel unsupervised machine learning framework that utilizes a graph-based deep autoencoder for multi-type and feature-based clustering based on heterogeneous drug data to identify safe, effective, affordable treatment options for COVID-19. We conclude that hierarchical clustering performed best for the given data out of the three unsupervised clustering algorithms. In addition to this, we also conclude  that GAE outperformed VGAE for learning latent embeddings, possibly because of a simpler architecture. Furthermore, this study lists the top 15 drugs that can be repurposed for COVID-19 treatment. These candidates have been selected on the basis of various non-linear relationships between different drug properties, such as biological properties, PK/PD, and assay data  which were captured by our machine learning framework. However, the outcomes of this study only implicate potential leads for drug-repurposing and warrant extensive experimental studies for testing and validating the top screened molecules for anti-COVID efficacy. Our machine-learning framework can be adapted for other drug repurposing studies in the future. 

\section*{Code and Data}

Code for the framework is available on Github\footnote{https://github.com/ai-covariants/drug-repurposing-covid19}. Note that the data is restricted. Please email the corresponding author for the dataset. 

\section*{Acknowledgement}

We would like to thank Dr Sankaranarayanan Murugesan, Muzaffar-Ur-Rehman and Hardik Ashish Jain for their crucial inputs in the dataset preparation stage of the study.  We would like to express our heartfelt gratitude to  Birla Institute of Technology and Science (BITS) Pilani for partially funding the project. This research was kindly funded by the National Health and Medical Research Council - Medical Research Future Fund (MRFF) (Grant Number: MRF2009092) administered by UNSW Sydney. International Programmes and Collaboration Division at BITS Pilani awarded a merit-based scholarship for this research.  We would also like to thank Prof. Seshadri Vasan from the University of York.

\section*{Appendix}
 
 Figure \ref{fig:DDR1} presents the graphical representation of the four DDR1 matrices which were obtained after Tier-1 clustering. 

 \begin{figure*}
 \centering
 \begin{subfigure}{0.475\textwidth}
     \includegraphics[width=\textwidth]{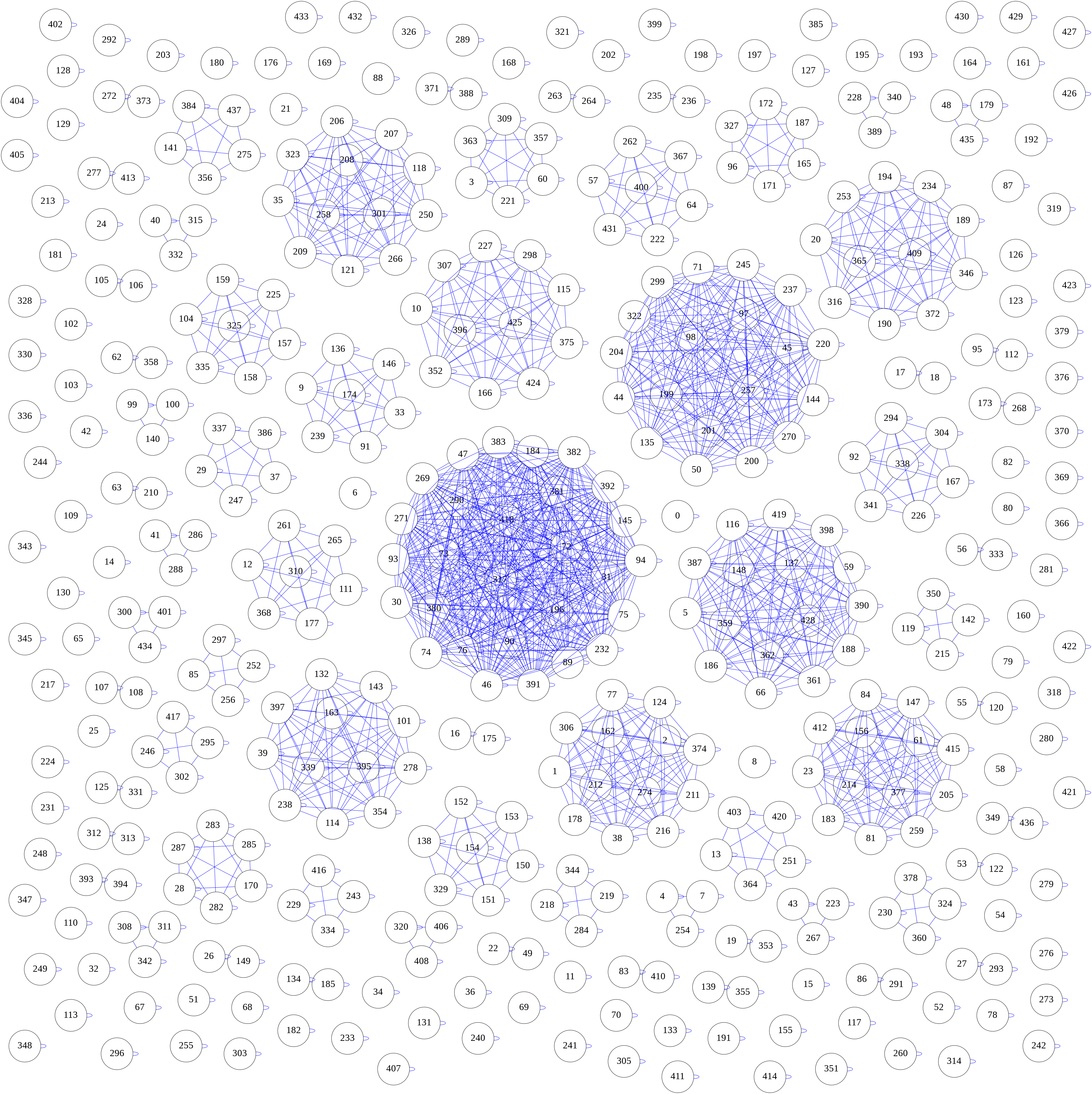}
     \caption{DDR1 - Drug-Indication}
 
 \end{subfigure}
 \hfill
 \begin{subfigure}{0.475\textwidth}
     \includegraphics[width=\textwidth]{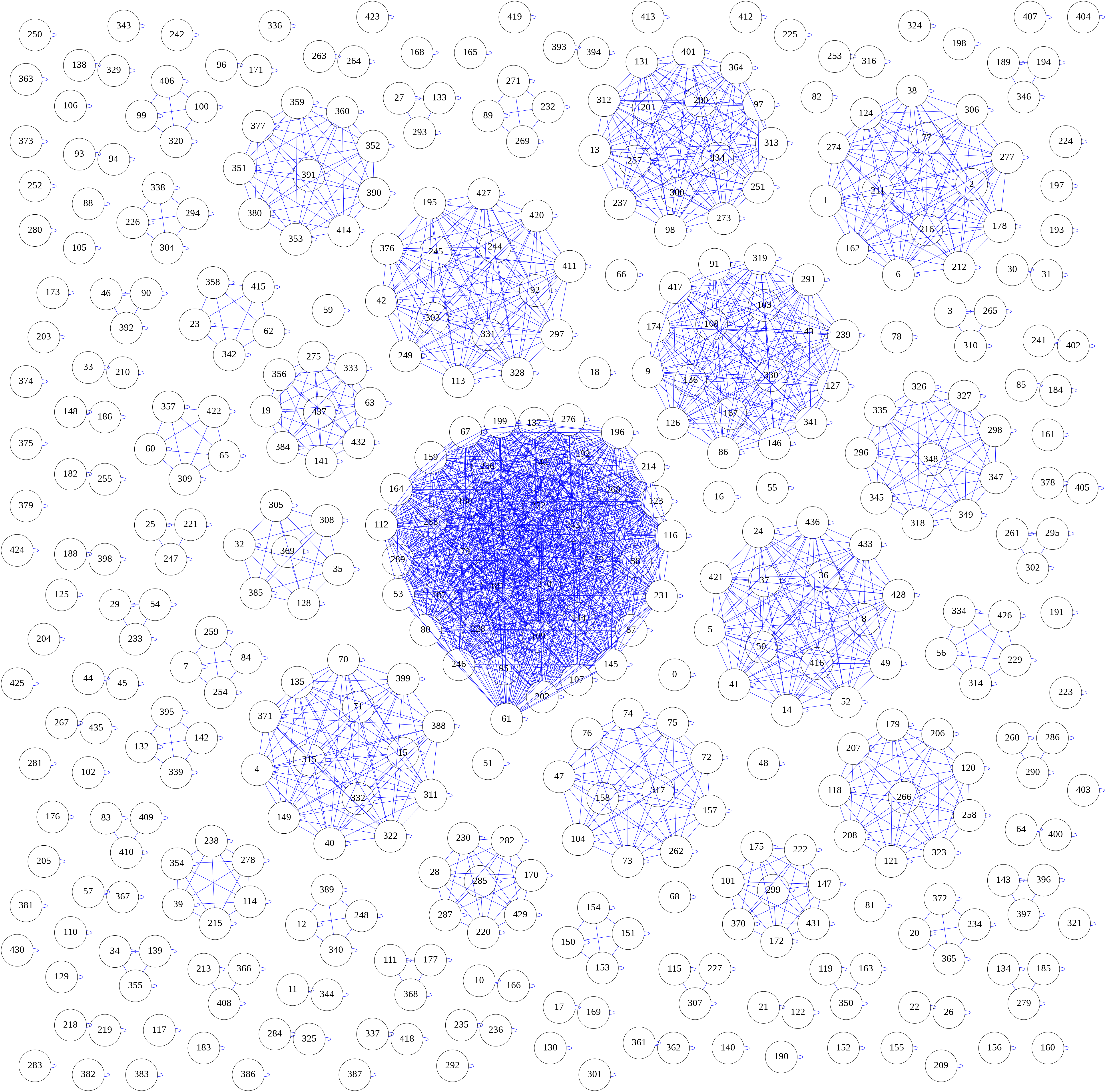}
     \caption{DDR1 - Drug-MOA}
 
 \end{subfigure}
 \hfill
 \vskip\baselineskip
 \begin{subfigure}{0.475\textwidth}
     \includegraphics[width=\textwidth]{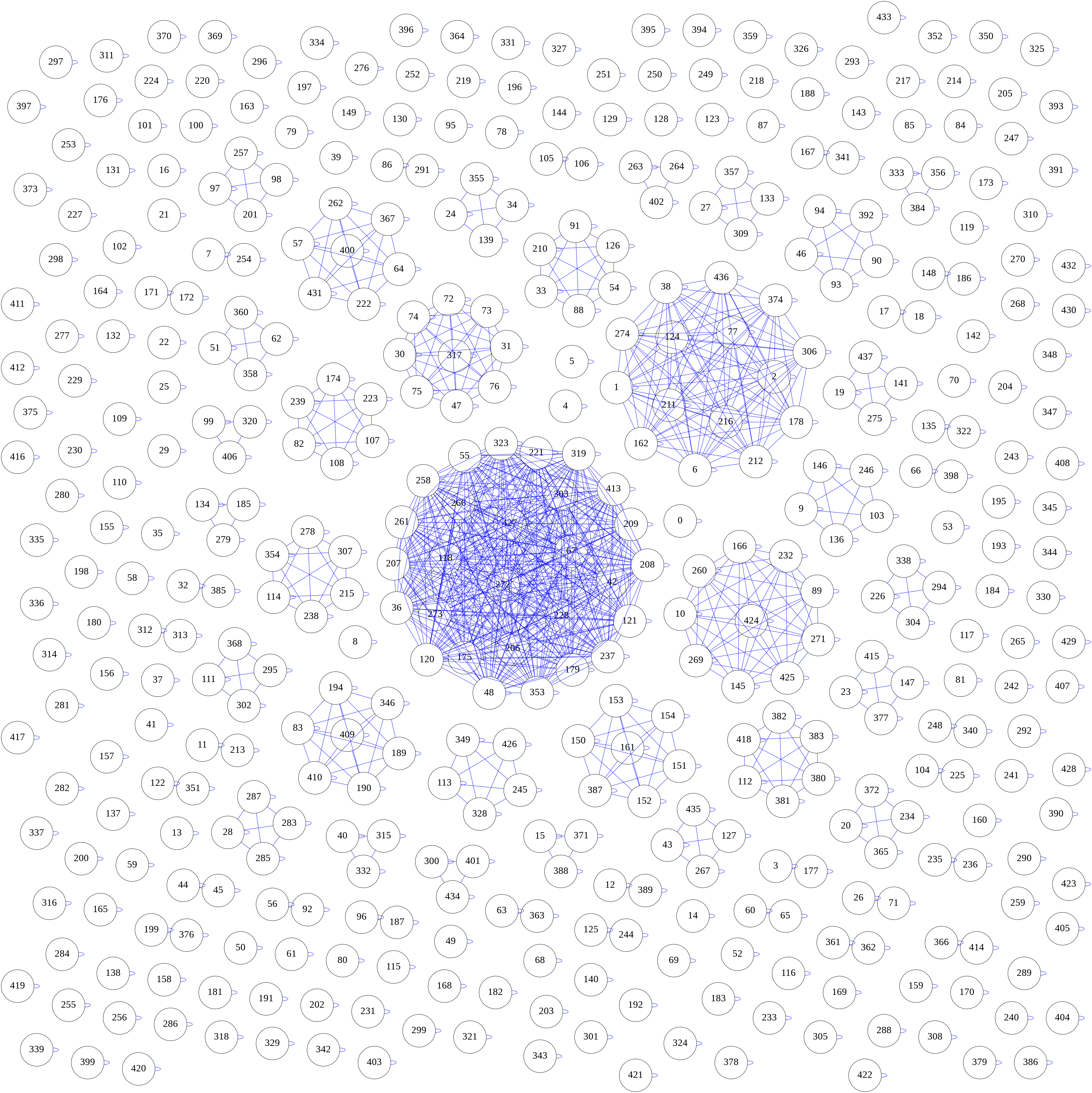}
    \caption{DDR1 - Drug-Pathway}
 
 \end{subfigure}
 \hfill
 \begin{subfigure}{0.475\textwidth}
     \includegraphics[width=\textwidth]{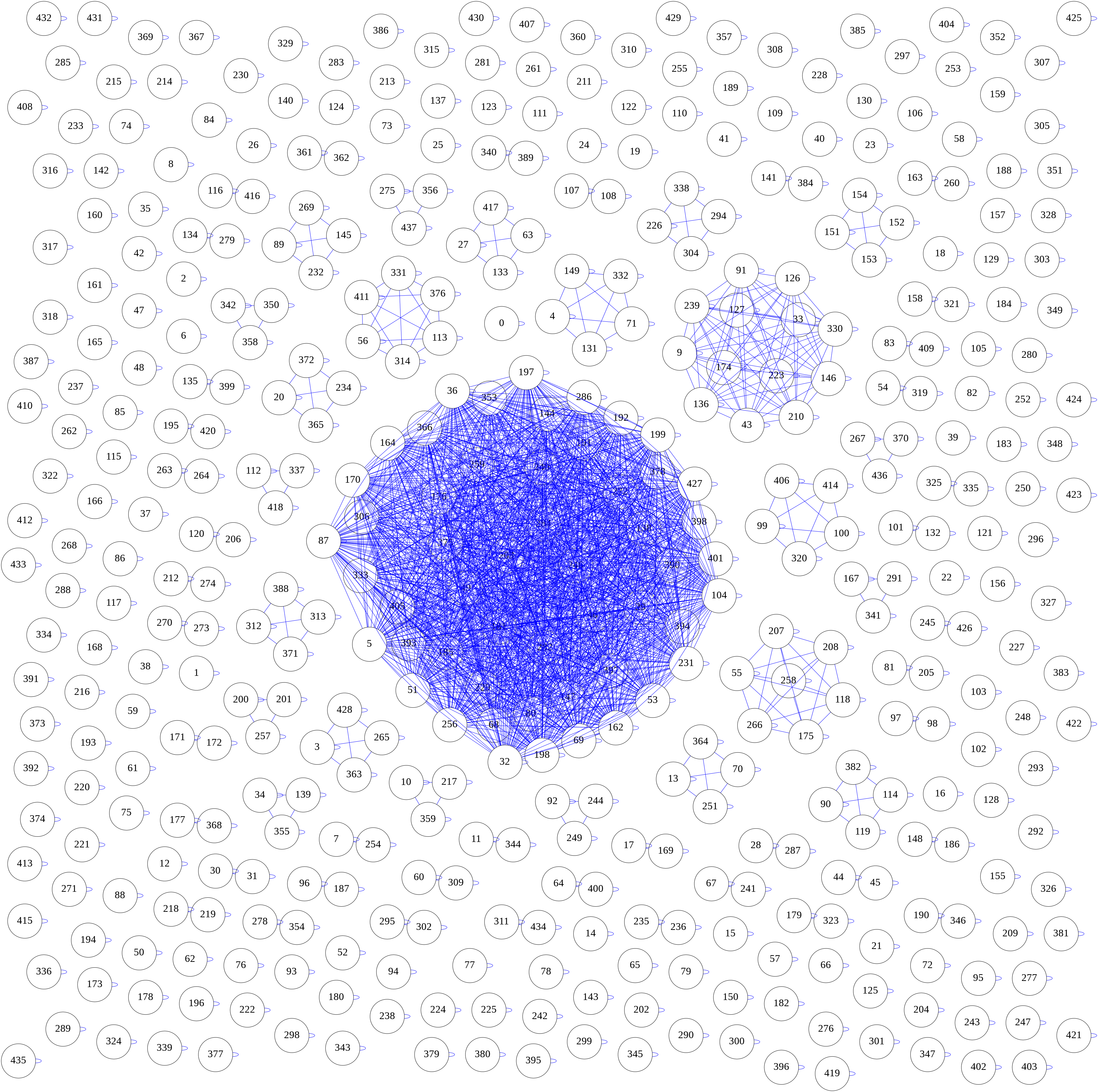}
     \caption{DDR1 - Drug-Target}
 
 \end{subfigure}
 \label{fig:DDR1}
 \caption{Graphical representation of DDR1 matrices formed after applying hierarchical clustering on individual textual features which are then combined to form final DDR graph depicted by Figure \ref{fig:DDR-final}}
\label{fig:DDR1}
\end{figure*}

\bibliographystyle{model1-num-names}
\bibliography{sample}
 
\end{document}